\documentclass[10pt, conference]{IEEEtran}

\IEEEoverridecommandlockouts                              

\usepackage{amsmath} 
\usepackage{amssymb}  
\usepackage{graphicx}

\title{Grammarization-Based Grasping with Deep Multi-Autoencoder Latent Space Exploration by Reinforcement Learning Agent}
\author{Leonidas Askianakis\\Technical University of Munich, Munich School of Engineering, \\ Department of Aerospace and Geodesy, Munich, Germany\\ \texttt{leonidas.askianakis@tum.de}}

\begin{document}

\maketitle
\thispagestyle{empty}
\pagestyle{empty}

\begin{abstract}

Grasping by a robot in unstructured environments is deemed a critical challenge because of the requirement for effective adaptation to a wide variation in object geometries, material properties, and other environmental factors. In this paper, we propose a novel framework for robotic grasping based on the idea of compressing high-dimensional target and gripper features in a common latent space using a set of autoencoders. Our approach simplifies grasping by using three autoencoders dedicated to the target, the gripper, and a third one that fuses their latent representations. This allows the RL agent to achieve higher learning rates at the initial stages of exploration of a new environment, as well as at non-zero shot grasp attempts. The agent explores the latent space of the third autoencoder for better quality grasp without explicit reconstruction of objects. By implementing the PoWER algorithm into the RL training process, updates on the agent's policy will be made through the perturbation in the reward-weighted latent space. The successful exploration efficiently constrains both position and pose integrity for feasible executions of grasps. We evaluate our system on a diverse set of objects, demonstrating the high success rate in grasping with minimum computational overhead. We found that approach enhances the adaptation of the RL agent by more than 35 \% in simulation experiments.

\end{abstract}

\section{Introduction}

Robotic grasping in unstructured and dynamic settings is still one of the significant challenges in robotics. Generalization and adaptation to new objects, environments, and tasks remain limited with a robot, even with recent progress in Reinforcement Learning and Deep Learning. In particular, grasping tasks are constituted by a multitude of factors, including object density distribution, mass, surface properties, environmental conditions, and more — many of which are either not explored or ignored in current methods \cite{bohg2014}. This results in lower adaptability and often suboptimal performance when robots encounter new or unexpected situations.
All these methods are often fully dependable on visual or haptic sensor-based inputs only, which cannot capture the complexity of real-world grasping scenarios \cite{billard2019}.
Moreover, most current grasping approaches ignore the dependency of task success on various environmental factors like humidity, temperature, lighting, and others. For instance, changes in friction between a gripper and an object will influence the force required for a stable grasp \cite{cheng2019}. Similarly, many other environmental factors affect the grasping algorithms' grasp quality and adaptation performance. Most well-performing strategies in a controlled environment fail when applied in dynamic and unpredictable environments \cite{jain2020}.

A limitation of the current RL-based grasping approaches is that they are unable to work adaptively or efficiently in new environments or with new objects for which they were not trained. Most current approaches require a large amount of training data or a great amount of time to fine-tune policies on novel tasks, preventing them from being employed in applications that profit most strongly from real-time performance in changing environments \cite{kroemer2021}. Further, this is complicated by high-dimensionality observation and action spaces for robotic manipulation tasks, making the learning process both slow and sample-inefficient \cite{chen2019}. In order to curb these challenges, there have been propositions for latent space representations; however, most do not encompass some critical physical properties of an object, such as mass, center of mass, or surface friction, which are fundamental for accomplishing a precise and stable grasp \cite{pande2019}.

\subsection{Grammarization of Grasping Components}

\noindent \textbf{Grammarization} is defined as the process of abstracting and encoding the physical properties and behaviors of multiple objects simultaneously into a set of computable metrics that, while preserving their essential information for the purpose that grammarization was performed. 

\vspace{10pt}

Mathematically, the grammarization process can be formalized as follows:

\vspace{10pt}

Given a feature vector \( f \in \mathbb{R}^n \) representing an object’s features and characteristics (extracted by any of the existing feature identification methodologies) such as mass, center of mass, geometrical features, and more, the grammarization function \( g: \mathbb{R}^n \to \mathbb{R}^m \) is defined such that:

\vspace{-0.5cm}

\[
 \boldsymbol{g}(\boldsymbol{f}) = \left( \theta_1(f_1, f_2, ..., f_n),  \theta_m(f_1, f_2, ..., f_n) \right) \in \mathbb{R}^{m}, m \leq n
\]

\noindent where \(\theta_i\) represents a specific mapping (grammar rule) with which the same features \( f_i \in \mathbb{R} \) are correlated in a different way. This process is surjective (but not not necessarily injective), meaning that every element in the lower-dimensional space corresponds to at least one element in the higher-dimensional space, ensuring the copmression of the critical properties of the object, and deliberately allowing different feature representations \( \boldsymbol{f} \in \mathbb{R}^n \) and \(\boldsymbol{f'} \in \mathbb{R}^n \) to lead to similar or even the same grammarized representations (\(\theta_i(f_1, f_2, ..., f_n) \approx \theta'_i(f_1, f_2, ..., f_n)\)).

\subsection{Main Contributions}

The main contributions of our work can be summarized as follows: 

\begin{itemize}
    \item \textbf{Grammarization of Gripper and Target Object}: We compress the high-dimensional properties of the gripper and target object into a common latent space using separate autoencoders, capturing both geometrical and physical parameters, providing a more comprehensive understanding of the manipulation scenario.
    
    \item \textbf{Reinforcement Learning in Latent Space}: We introduce a reinforcement learning framework where the agent operates in the compressed latent space of the gripper-target correlation, enabling faster learning and adaptation by exploring a lower-dimensional yet highly informative space \cite{kim2020}. 
    
    \item \textbf{Environmental and Physical Integration}: Our framework integrates physical parameters (e.g., mass, friction, moments of inertia) and environmental factors (e.g., humidity, temperature, solar irradiation), broadening its applicability to dynamic real-world scenarios \cite{cheng2019, jain2020}.
\end{itemize}

\section{RELATED WORK}

Robotic grasping has made significant advances using deep reinforcement learning (RL) and latent space representations, especially in structured environments. However, challenges persist in unstructured settings where generalization to new objects and real-time decision-making are critical. Traditional methods like CNNs and DMPs succeed in controlled environments but often neglect crucial object properties and environmental factors essential for dynamic grasping \cite{chen2021_drl}, \cite{wang2020}.

Previous works have explored improving grasping strategies through RL in reduced latent spaces, but they often focus primarily on visual features, ignoring key physical properties like mass and surface friction \cite{popov2017}. For instance, Popov et al.'s use of DDPG in dexterous manipulation, and Joshi et al.'s deep Q-learning for robotic grasping, heavily rely on visual inputs without integrating essential physical characteristics \cite{joshi2020}.

Autoencoders (AEs) have been applied in reducing task complexity, as shown by Rezaei-Shoshtari et al. and Zhao et al., who focus on visual data for dynamic tasks and 3D pose estimation, respectively. However, these approaches do not address non-visual properties or environmental conditions \cite{rezaei2020}, \cite{zhao2023}. PoseRBPF integrates 3D pose and vector information but does not fully explore latent spaces incorporating both visual and non-visual features for adaptive grasping \cite{deng2019}. Chen et al.’s structure-preserving AEs preserve geometric relations but do not extend to encoding physical parameters \cite{chen2021_vae}.

\section{PROPOSED ARCHITECTURE}

In the quest to simplify exploration spaces for reinforcement learning (RL) agents and face the curse of dimensionality, our architecture incorporates three autoencoders (AE), each focusing on different aspects of the grasping problem: target grammarization (AE\(_1\)), gripper grammarization (AE\(_2\)), and joint integration through AE\(_3\). 

\subsection{Target Grammarization: }

The AE\(_1\) autoencoder receives voxelized representations of the target object derived from CAD models, with future adaptability to integrate 6D pose estimations from computer vision systems. Voxelization is widely used in grasp planning due to its efficiency in representing complex 3D geometries, allowing the system to capture volumetric and surface details crucial for manipulation \cite{mousavian2019}. 

For the encoder, we adopted a 3D convolutional neural network (CNN) architecture inspired by FeatureNet, an approach for machining feature recognition from 3D CAD models whose dataset we also used for the training of the target object grammarization autoencoder \cite{FeatureNet}. The encoder part of AE\(_1\) uses 3D CNN architecture with additional input parameters of mass, principal moments of intertia, and surface friction coefficient (estimated for 3D printed objects with 0.1 print layer accuracy made out of PLA). The filter sizes were 3x3x3 followed by pooling layers, progressively reducing the input voxelized grid into a compressed latent representation. The network consists of three convolutional layers, each followed by ReLU activations and max-pooling operations to reduce spatial dimensions while retaining key geometric information. The final layers of the encoder map contain the extracted features to a lower-dimensional latent space on which the key physical properties of the target object are also added. The decoder part of AE\(_1\) follows the necessary inversion layers to achieve a reconstruction of the partitioned entry vector representing the shape of the target and its physical properties, all together achieving a quite accurate representation of its state.

\begin{figure}[h]  
\centering
\includegraphics[width=0.5\columnwidth]{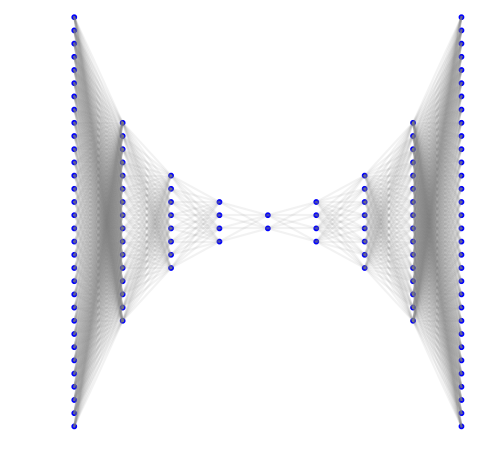}  
\caption{Qualitative representation of \(AE_1\)}
\label{fig:your_label}
\end{figure}

The latent space produced by AE\(_1\) captures a compressed representation of the target object, retaining key features such as shape, surface contours, and physical-related properties. Let $ X \in \mathbb{R}^{n \times n \times n} $ represent the voxelized object shape, where $ n $ is the voxel grid size (e.g., 32x32x32). The encoder maps this high-dimensional input $ X $ into a lower-dimensional latent vector $ z \in \mathbb{R}^m $, where $ m \ll n $. This latent space is a compressed, yet highly informative, representation of the object. The decoder reconstructs the voxel grid back from the latent space, represented as $ \hat{X} = D_\phi(z) $. The learning objective of AE\(_1\) is to minimize the following loss function:
$$
\mathcal{L} = \|X - \hat{X}\|^2 \eqno{(1)}
$$
This loss function consists of a reconstruction error (mean squared error) between the input and the reconstructed output.

\subsection{Grammarization of the Gripper - AE\(_2\)}

For the grammarization of the gripper we follow a similar architecture, on which the information of the pose information with respect to the target frame had to also be included.

\textbf{Input Representation:} The input to \textbf{AE\(_2\)} is twofold: A voxelized representation of the gripper's structure and a \textbf{pose vector} representing the gripper's initial position and orientation with respect to the target frame. Let the voxelized representation of the gripper be denoted as \( G(x, y, z) \), where \( x, y, z \) are the spatial coordinates of each voxel. The pose vector is denoted as \( \mathbf{p} \), which consists of two parts: a positional vector \( \mathbf{r} \in \mathbb{R}^3 \) and a quaternion representation of the orientation \( \mathbf{q} \in \mathbb{R}^4 \). Thus, the complete input can be defined as:
$$
\mathbf{I}_G = \{ G(x, y, z), \mathbf{r}, \mathbf{q} \} \eqno{(2)}
$$

Where \( G(x, y, z) \) is a 3D tensor, \( \mathbf{r} = [r_x, r_y, r_z] \), and \( \mathbf{q} = [q_w, q_x, q_y, q_z] \). This encapsulates both the gripper's geometry and its spatial relation to the target.

\textbf{Encoding the Gripper and Pose:} 
Like in AE\(_1\), the The CNN applies a series of 3D convolutional layers to extract key features from the gripper’s geometry, such as the shape, size, surface properties, and contact area. Let the feature extraction be represented as:
$$
\mathbf{F}_G = f_{\text{CNN}}(G(x, y, z)) \eqno{(3)}
$$

Where \( f_{\text{CNN}} \) denotes the series of 3D convolutional layers, and \( \mathbf{F}_G \) is the extracted feature representation of the gripper’s structure.

The \textbf{pose vector} \( \mathbf{p} = [\mathbf{r}, \mathbf{q}] \) is concatenated with the output of the CNN to encode both spatial and geometrical information simultaneously. Instead of processing the pose vector through convolutions, we feed it into fully connected layers to compress it into a latent representation:
$$
\mathbf{F}_p = f_{\text{FC}}(\mathbf{p}) \eqno{(4)}
$$

Where \( f_{\text{FC}} \) represents the fully connected layers used to compress the pose vector.

\begin{figure}[h]  
\centering
\includegraphics[width=0.5\columnwidth]{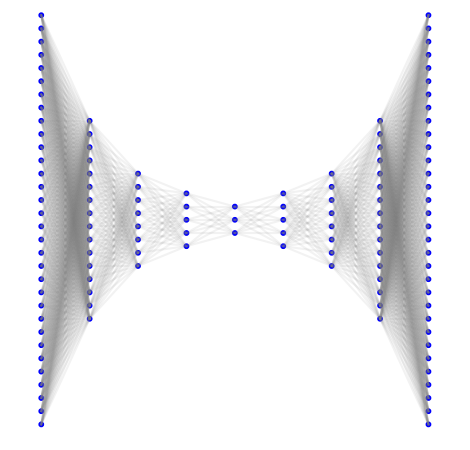}  
\caption{Qualitative representation of \(AE_2\)}
\label{fig:your_label}
\end{figure}

The encoded gripper structure \( \mathbf{F}_G \) and pose vector \( \mathbf{F}_p \) are then concatenated to form the complete latent representation of the gripper:
$$
\mathbf{z}_G = [\mathbf{F}_G, \mathbf{F}_p] \eqno{(5)}
$$

This latent vector \( \mathbf{z}_G \) is the output of \textbf{AE\(_2\)}’s encoder, which represents both the structural and pose characteristics of the gripper in a compact form. This vector will later be used as part of the input to \textbf{AE\(_3\)}. Note that the optimal latent space has been found to be optimal in higher dimensions than \textbf{AE\(_1\)} because of the encoding of the extra not so correletable parameters of the pose vector.

\textbf{Latent Space and Reconstruction:} 
The latent space of \textbf{AE\(_2\)} is designed to allow both the reconstruction of the gripper’s geometry and its pose after the perturbations that occur due to \textbf{AE\(_3\)}. The latent vector \( \mathbf{z}_G \) is passed into the decoder, which reconstructs both the voxelized gripper structure and the pose vector.

Let the decoder's function be represented as:
$$
\hat{G}(x, y, z), \hat{\mathbf{p}} = f_{\text{dec}}(\mathbf{z}_G) \eqno{(6)}
$$

Where \( f_{\text{dec}} \) is the decoder function, \( \hat{G}(x, y, z) \) is the reconstructed voxelized gripper, and \( \hat{\mathbf{p}} = [\hat{\mathbf{r}}, \hat{\mathbf{q}}] \) is the reconstructed pose vector.

The loss function for \textbf{AE\(_2\)} includes a \textbf{reconstruction loss} for both the geometry and the pose:
$$
\mathcal{L}_{\text{recon}} = \| G(x, y, z) - \hat{G}(x, y, z) \|_2^2 + \| \mathbf{p} - \hat{\mathbf{p}} \|_2^2 \eqno{(7)}
$$

The total loss for \textbf{AE\(_2\)} is the reconstruction loss. No Gaussian noise or KL-divergence regularization is applied.

\subsection{Autoencoder 3: Grammarization of the Combined Gripper-Target Latent Space}
\vspace{0.2cm}

The goal of \textbf{AE\(_3\)} is to integrate the latent spaces from \textbf{AE\(_1\)} (target grammarization) and \textbf{AE\(_2\)} (gripper grammarization) into a unified latent space. This combined space simplifies the task of the reinforcement learning (RL) agent by reducing the complexity of the exploration space while maintaining the essential features required for accurate and adaptable grasping.

The key challenge is to effectively combine the two separate latent spaces (\( \mathbf{z}_T \) for the target, and \( \mathbf{z}_G \) for the gripper) into a single, compressed latent representation. Additionally, constraints must be imposed to ensure the integrity of each latent space after encoding and decoding, allowing accurate reconstructions of both the gripper and the target.

\subsubsection{Input Representation and Encoding}

The input to \textbf{AE\(_3\)} is the concatenation of the latent spaces from \textbf{AE\(_1\)} and \textbf{AE\(_2\)} on which it has also been trained on. Let the latent vector for the target be represented by \( \mathbf{z}_T \in \mathbb{R}^{m_T} \), and the latent vector for the gripper be represented by \( \mathbf{z}_G \in \mathbb{R}^{m_G} \). The concatenated latent vector \( \mathbf{z}_{GT} \in \mathbb{R}^{m_T + m_G} \) is then formed as:
$$
\mathbf{z}_{GT} = [\mathbf{z}_T, \mathbf{z}_G] \eqno{(8)}
$$

The encoder of \textbf{AE\(_3\)} compresses this combined latent vector into a more compact latent representation \( \mathbf{z}_C \in \mathbb{R}^{m_C} \), where \( m_C < (m_T + m_G) \). This compression reduces the dimensionality of the search space for the RL agent, making learning more efficient, although due to the less correlatable characteristics between the optimal latent space representation was achieved in relatively high dimensions - as expected. The encoder function \( E_3 \) maps the concatenated latent vector to the compressed latent space as:
$$
\mathbf{z}_C = E_3(\mathbf{z}_{GT}) \eqno{(9)}
$$
Where \( E_3 \) represents the neural network that encodes the combined latent space.

\subsubsection{Latent Space Constraints and Structure}

The key innovation in \textbf{AE\(_3\)} is the application of \textbf{constraints} on the latent spaces, ensuring that certain components of the concatenated latent space \( \mathbf{z}_{GT} \) retain their position during encoding and decoding. Specifically, the following constraints are applied:

1. \textbf{Positional Constraints:} The latent space \( \mathbf{z}_G \) representing the gripper must remain in a fixed position within \( \mathbf{z}_{GT} \), both before and after encoding and decoding by \textbf{AE\(_3\)} in a similar way to the constraints that have been applied to the position of the output of \textbf{AE\(_2\)} to account for the location of the pose parameters. Let \( z_G[i] \) represent the \(i\)-th entry of \( \mathbf{z}_G \), then the positional constraint ensures that:
$$
z_G[i] \longrightarrow \hat{z}_G[i] \quad \text{for} \quad i \in \{1, 2, \dots, m_G\} \eqno{(10)}
$$
Where \( \hat{z}_G[i] \) represents the decoded latent vector of the gripper. This ensures that gripper-specific features are not distorted during the encoding process.

2. \textbf{Target Integrity:} Similarly, the latent space \( \mathbf{z}_T \) for the target must retain its integrity during encoding and decoding. This constraint ensures that the target features are decoded accurately, and there is no cross-mixing of target and gripper features:
$$
z_T[i] \longrightarrow \hat{z}_T[i] \quad \text{for} \quad i \in \{1, 2, \dots, m_T\} \eqno{(11)}
$$
Where \( \hat{z}_T[i] \) represents the decoded latent vector of the target. In figure 3, the layers \(l_4\) and \(l_{10}\) are representing the latent vector, which is the input and output layer of the \ \textbf{AE\(_3\)}.

\begin{figure}[h]  
\centering
\includegraphics[width=\columnwidth]{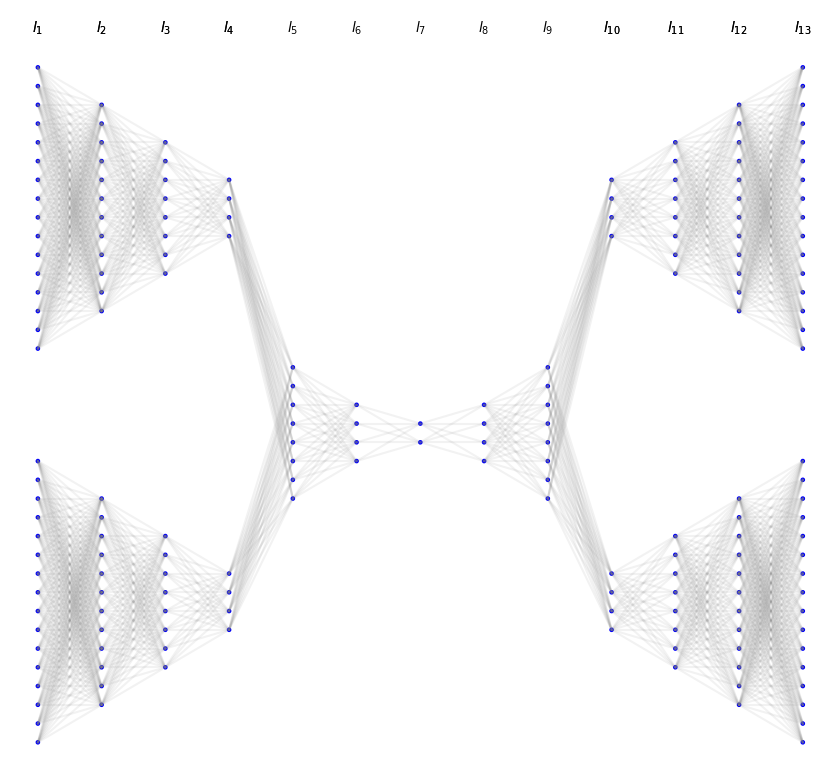}  
\caption{Qualitative representation of \(AE_3\)}
\label{fig:your_label}
\end{figure}

3. \textbf{Pose and Grasp Information:} The pose information for the gripper, included as part of \( \mathbf{z}_G \), must also be accurately reconstructed. The pose vector \( \mathbf{p} = [\mathbf{r}, \mathbf{q}] \) consists of the position and orientation of the gripper with respect to the target, and any change in these values should be reflected correctly in the output. The constraint on the pose vector ensures that:
$$
\mathbf{p} = [\mathbf{r}, \mathbf{q}] \longrightarrow \hat{\mathbf{p}} = [\hat{\mathbf{r}}, \hat{\mathbf{q}}] \eqno{(12)}
$$
Where \( \hat{\mathbf{r}} \) and \( \hat{\mathbf{q}} \) represent the reconstructed position and quaternion values for the gripper.

\subsubsection{Decoding and Reconstruction}

The decoder of \textbf{AE\(_3\)} maps the compressed latent vector \( \mathbf{z}_C \) back to the concatenated latent space \( \hat{\mathbf{z}}_{GT} \), which is then split into the latent spaces \( \hat{\mathbf{z}}_G \) and \( \hat{\mathbf{z}}_T \) for the gripper and the target, respectively:
$$
\hat{\mathbf{z}}_{GT} = D_3(\mathbf{z}_C) \eqno{(13)}
$$
Where \( D_3 \) represents the decoder network of \textbf{AE\(_3\)}. The latent spaces are then passed back to the decoders of \textbf{AE\(_1\)} and \textbf{AE\(_2\)} for the final reconstruction:
$$
\hat{G}(x, y, z), \hat{\mathbf{p}} = D_2(\hat{\mathbf{z}}_G) \quad \text{and} \quad \hat{T}(x, y, z) = D_1(\hat{\mathbf{z}}_T) \eqno{(14)}
$$
Where \( \hat{G}(x, y, z) \) represents the reconstructed gripper, \( \hat{\mathbf{p}} \) is the reconstructed pose, and \( \hat{T}(x, y, z) \) represents the reconstructed target object.

\subsubsection{Loss Function}

The total loss for \textbf{AE\(_3\)} is designed to ensure accurate reconstruction of both the gripper and target latent spaces, as well as the combined latent space. It is composed of the following components:

1. \textbf{Reconstruction Loss:} This measures the error between the original latent space \( \mathbf{z}_{GT} \) and the reconstructed latent space \( \hat{\mathbf{z}}_{GT} \). The reconstruction loss for the combined latent space is:
$$
\mathcal{L}_{\text{recon}} = \| \mathbf{z}_{GT} - \hat{\mathbf{z}}_{GT} \|_2^2 \eqno{(15)}
$$

2. \textbf{Gripper and Target Losses:} To ensure that the individual latent spaces \( \mathbf{z}_G \) and \( \mathbf{z}_T \) are accurately reconstructed, additional reconstruction losses are imposed:
$$
\mathcal{L}_{G} = \| \mathbf{z}_G - \hat{\mathbf{z}}_G \|_2^2 \quad \text{and} \quad \mathcal{L}_{T} = \| \mathbf{z}_T - \hat{\mathbf{z}}_T \|_2^2 \eqno{(16)}
$$

3. \textbf{Pose Reconstruction Loss:} The error in reconstructing the gripper's pose vector \( \mathbf{p} = [\mathbf{r}, \mathbf{q}] \) is also measured using a separate loss term:
$$
\mathcal{L}_{\text{pose}} = \| \mathbf{p} - \hat{\mathbf{p}} \|_2^2 \eqno{(17)}
$$

The total loss for \textbf{AE\(_3\)} is the weighted sum of these components:
$$
\mathcal{L}_{\text{AE3}} = \mathcal{L}_{\text{recon}} + \alpha (\mathcal{L}_{G} + \mathcal{L}_{T}) + \beta \mathcal{L}_{\text{pose}} \eqno{(18)}
$$
Where \( \alpha \) and \( \beta \) are regularization parameters that control the relative importance of the gripper-target reconstruction and pose accuracy.

\subsection{Reinforcement Learning in Latent Space}

The latent space \(\mathbf{z}_C\) of AE\(_3\) is formed from the concatenated latent spaces of AE\(_1\) and AE\(_2\):
$$
\mathbf{z}_{GT} = [\mathbf{z}_T, \mathbf{z}_G] \eqno{(19)}
$$
where \(\mathbf{z}_T \in \mathbb{R}^{m_T}\) is the latent encoding of the target, and \(\mathbf{z}_G \in \mathbb{R}^{m_G}\) is the latent encoding of the gripper. The encoder of AE\(_3\), \(E_3\), compresses this combined latent space \(\mathbf{z}_{GT}\) into a lower-dimensional latent space:
$$
\mathbf{z}_C = E_3(\mathbf{z}_{GT}) \eqno{(20)}
$$
where \(\mathbf{z}_C \in \mathbb{R}^k\), and \(k < m_T + m_G\), ensuring dimensionality reduction. The RL agent explores this latent space \(\mathbf{z}_C\), applying perturbations to discover the best actions for grasping. The agent’s action \(a\) is represented as a perturbation \(\delta\) applied to \(\mathbf{z}_C\):
$$
\mathbf{z}_C' = \mathbf{z}_C + \delta \eqno{(21)}
$$
The goal is to find the optimal perturbation \(\delta^*\) that maximizes the expected grasping reward \(R\):
$$
\delta^* = \arg \max_\delta \mathbb{E}[R(\mathbf{z}_C + \delta)] \eqno{(22)}
$$
The perturbation \(\delta\) is drawn from a distribution that evolves over time as the RL agent learns. The RL agent updates its policy by weighting the perturbations based on the observed reward.

The reward function \(R\) guides the learning of the RL agent. At the initial stage, the reward function prioritizes grasping quality over reconstruction accuracy, as the focus is on optimizing the interaction between the gripper and the target. The reward function is defined as:
$$
R = f(\text{Grasp Quality}) - \alpha \|\hat{\mathbf{z}}_T - \mathbf{z}_T\|^2 - \beta \|\hat{\mathbf{z}}_G - \mathbf{z}_G\|^2 \eqno{(23)}
$$
where:
- \(f(\text{Grasp Quality})\) measures the success of the grasp, considering factors such as whether the object was lifted, the stability of the grasp, and force exertion.
- \(\hat{\mathbf{z}}_T\) and \(\hat{\mathbf{z}}_G\) are the reconstructed latent encodings for the target and gripper, respectively.
- \(\alpha\) and \(\beta\) are small weights to ensure that reconstruction errors are not heavily penalized at this stage, keeping the focus on grasping.

The PoWER (Policy learning by Weighting Exploration with the Returns) method is employed for updating the policy of the RL agent \cite{Kober2011}. The PoWER algorithm is particularly suitable for tasks like robot grasping because it effectively balances exploration and exploitation, updating policies based on reward-weighted perturbations. The policy \(\pi_\theta(a | \mathbf{z}_C)\), where \(\theta\) represents the policy parameters and \(a\) represents the perturbations, is updated as follows:
$$
\theta_{new} = \theta_{old} + \eta \sum_{i=1}^N \frac{R_i}{\sum_j R_j} (\theta_i - \theta_{old}) \eqno{(24)}
$$
where:
- \(R_i\) is the reward for trajectory \(i\),
- \(N\) is the number of sampled trajectories, and
- \(\eta\) is the learning rate.
The policy is updated iteratively, with the agent gradually favoring actions (perturbations) that lead to higher rewards, ensuring convergence toward an optimal grasping strategy.

Two critical constraints are applied during the exploration of the latent space:
1. \textbf{Positional Constraints:} The latent representations of the target and gripper, \(\mathbf{z}_T\) and \(\mathbf{z}_G\), must retain their positional integrity within the combined latent space. 
2. \textbf{Pose Integrity Constraints:} The pose vector, encoded as part of \(\mathbf{z}_G\), must be accurately reconstructed to preserve the gripper’s ability to execute a successful grasp. Pose information is encoded as quaternions in the latent space, ensuring orientation and position are preserved during latent space exploration. The constrained optimization problem can be expressed as:
$$
\min_{\delta} \|\hat{\mathbf{z}}_T - \mathbf{z}_T\|^2 + \|\hat{\mathbf{z}}_G - \mathbf{z}_G\|^2 \quad  \eqno{(25)}
$$

Once the RL agent has applied a perturbation to the latent space \(\mathbf{z}_C\), the decoder of AE\(_3\) reconstructs the combined latent space \(\mathbf{z}_{GT}\):
$$
\hat{\mathbf{z}}_{GT} = D_3(\mathbf{z}_C') \eqno{(26)}
$$
where \(\mathbf{z}_C'\) is the perturbed latent space after exploration. The decoders of AE\(_1\) and AE\(_2\) then reconstruct the target and gripper:
$$
\hat{T} = D_1(\hat{\mathbf{z}}_T), \quad \hat{G}, \hat{p} = D_2(\hat{\mathbf{z}}_G) \eqno{(27)}
$$
At this stage, the RL agent does not focus on achieving perfect reconstructions but instead prioritizes optimizing the grasping policy. Reconstruction becomes a higher priority in future phases.

The RL training is terminated once the agent achieves a grasp success rate above a predefined threshold, \(R_{success}\). The criteria are based on the agent's ability to consistently lift and stabilize the target across multiple episodes. The force applied and the stability of the grasp (evaluated by \(f(\text{Grasp Quality})\)) are key metrics. Mathematically, the training stops when the success rate over the past \(M\) episodes exceeds the threshold:
$$
\frac{1}{M} \sum_{i=1}^M \mathbb{E}[R_i] \geq R_{success} \eqno{(28)}
$$
In scenarios where reconstruction of the gripper is considered (future work), the training will include a second phase where the agent optimizes for both grasp success and object reconstruction accuracy.

\section{Experimental Setup and Results}

\subsection{Dataset Creation}

\textbf{Target Object Dataset (AE\(_1\)):} The dataset for \textbf{AE\(_1\)} consisted of 24,000 objects constructed by 10cm from which assumed manufacturing proceedures have been applied according to \cite{FeatureNet}, voxelized into a 16x16x16 grid, derived from CAD models. Each object was randomly perturbed, including rotation, scaling, and translation, to generate variability and all together generated 140000 samples. The voxelized representation was fed into \textbf{AE\(_1\)} for training.

\textbf{Gripper Dataset (AE\(_2\)):} The gripper dataset included 14,000 unique fingertip designs generated by modifying modular fingertip primitives for a Franka Emika hand. Each fingertip was voxelized and grammatically compressed. The creation of the dataset was achieved by perturbing the point cloud of the contact surface of the fingertip and generating new grippers with different properties. Additionally, pose information (quaternions) was integrated into the input data, representing the initial gripper-target alignment.

Both datasets were used to train the respective autoencoders, and then representative samples of both were selected and integrated into a Gazebo environment to simulate robotic grasping tasks for the RL agent. Once \textbf{AE\(_1\)} and \textbf{AE\(_2\)} were trained, their latent spaces were combined and fed into \textbf{AE\(_3\)} for training and further dimensionality reduction.

\subsection{Results}

The autoencoders were evaluated based on their reconstruction accuracy, defined as the percentage of correctly reconstructed elements (voxels for \textbf{AE\(_1\)}, features for \textbf{AE\(_2\)}, and the combined space for \textbf{AE\(_3\)}). The RL agent was then tasked with optimizing grasping performance in the latent space of \textbf{AE\(_3\)} using the PoWER algorithm.

\textbf{AE\(_1\)} achieved a reconstruction accuracy of more than 90\% on the 140,000 target object dataset. \textbf{AE\(_2\)} obtained more than 85\% accuracy on the gripper dataset including pose vector reconstruction. \textbf{AE\(_3\)}, which compresses both latent spaces, achieved an overall accuracy of more than 71\% for latent spaces reconstruction with positional constraints. 

For the RL agent, the learning rate was defined by the time required for the agent to reach an 80\% grasp success rate. Using the latent space of \textbf{AE\(_3\)}, the RL agent demonstrated a 35.8\% faster adaptation rate between when a gripper or a target has been altered, compared to the baseline methodology exploring on the observable envrionment on which the same alternation on gripper and/or targets happened and the RL agent was expected to re-increase the successful grasp rates.

\begin{table}[h]
\centering
\caption{Autoencoder Performance and RL Agent Adaptation Results}
\begin{tabular}{|c|c|c|}
\hline
\textbf{Model} & \textbf{Accuracy (\%)} & \textbf{Notes} \\
\hline
\textbf{AE\(_1\)} & 90.52 & 16x16x16 (140,000 samples) \\
\textbf{AE\(_2\)} & 85.23 & Gripper dataset (14,000 samples) \\
\textbf{AE\(_3\)} & 71.16 & Combined latent space  \\
\hline
\textbf{RL Agent} & 35.8\% improvement & Faster adaptation rate\\
\hline
\end{tabular}
\end{table}

The training process for the RL agent took place in a simulated Gazebo environment. Each trial began with the object from the \textbf{AE\(_1\)} dataset and the corresponding gripper from the \textbf{AE\(_2\)} dataset placed in the environment. The RL agent explored the latent space of \textbf{AE\(_3\)}, learning how to manipulate the gripper effectively for successful grasping tasks. The rewards were based on the ability of the gripper to lift and stabilize the object.

The combination of pre-trained autoencoders and the RL exploration in the compressed latent space resulted in a significant improvement in adaptation rates, as demonstrated by the faster convergence times. Each episode in Gazebo was evaluated based on grasp success, stability, and applied force, ensuring a robust training environment for real-world applications.

\vspace{-6pt}

\section{DISCUSSION}

The experimental results validate the effectiveness of our autoencoder-based grammarization framework in simplifying the robotic grasping task. By compressing high-dimensional features of the target object and gripper, the reinforcement learning (RL) agent operates in a reduced search space, leading to faster adaptation and more efficient learning. Incorporating physical and environmental features such as mass, center of mass, and friction coefficients enhances the system’s adaptability, enabling more generalizable grasping strategies.

The key advantage of this approach is its ability to mitigate the curse of dimensionality while retaining essential information for high-quality grasps. This proves beneficial in high-dimensional manipulation tasks, where traditional RL methods are inefficient. Our framework allows faster policy convergence by operating in a compressed latent space.

The reconstruction accuracy of \textbf{AE\(_1\)} and \textbf{AE\(_2\)} was 90\% and 85\%, respectively, while \textbf{AE\(_3\)} achieved 79\%, indicating room for improvement in the combined latent space. Additionally, the PoWER algorithm enhanced RL training efficiency by 35\%, but future work could explore other methods like PPO or TRPO to improve robustness.

Lastly, our grammarization framework shows promise in non-zero-shot grasping tasks, where grasp attempts during the execution process could be afforded like in on-orbit and maritime robotics, where grasping conditions are unpredictable. Future research could also explore multi-finger grippers and dry adhesive gripper design optimization based on latent space reconstruction. Part of the future work includes the generation of the suitable gripper for a specific manipulation scenario, by the use of masked autoencoders, on which parts, or even the entire gripper are unknown to the grammarization framework, and the outputs of \(AE_3\) and successively \(A_2\) try to identify which gripper would fit the manipulation scenario best during exploration, according to the data during the training of \(AE_2\) and \(AE_3\).


\section*{ACKNOWLEDGMENT}

At this point, we would like to acknowledge the use of generative AI for the grammatical enhancement of parts of the text of this manuscript.


\section*{PREPRINT NOTICE}
This paper is a preprint of work submitted for possible publication at the IEEE International Conference on Robotics and Automation (ICRA) 2025.



\end{document}